\documentclass[runningheads]{llncs}
\usepackage[T1]{fontenc}
\usepackage{amsmath}
\usepackage{graphicx,verbatim}
\usepackage{multirow} 
\usepackage[table]{xcolor}
\usepackage{booktabs}
\usepackage{subcaption}
\begin{document}
\title{S2T-Unet: A Structure-to-Style Framework for Inter-Modality MRI Translation}
%

\author{Yichao Liu\inst{1} }  
\authorrunning{Anonymized Author et al.}
\institute{Heidelberg University, IWR, Germany }

\maketitle              
\begin{abstract}
Inter-modality MRI translation aims to synthesize missing MRI modalities from available acquisitions, reducing the need for additional scanning while preserving clinically relevant anatomical information. However, existing image translation methods often learn intensity mappings without explicitly separating modality-invariant structural information from modality-specific appearance, which may lead to structural information loss or unrealistic image details. In this work, we propose S2T-Unet, a structure-to-style framework that explicitly models these two aspects. Specifically, vector quantization is introduced at the lower-level bottleneck to encode modality-invariant structural information using a learned discrete codebook. At higher levels, a modality transformation module uses decoder features to condition and transform encoder representations toward the target modality, thereby recovering modality-specific intensity and contrast information. Experiments on the IXI multi-contrast MRI dataset across four translation tasks demonstrate that S2T-Unet is comparable or outperform with state-of-art method.

\keywords{MRI  \and Image translation \and Vector quantization.}

\end{abstract}
\section{Introduction}
Magnetic Resonance Imaging (MRI) is a cornerstone of modern clinical diagnosis, offering excellent soft-tissue contrast without ionizing radiation \cite{hussain2022modern}. A comprehensive MRI protocol typically acquires multiple complementary sequences, such as T1-weighted, T2-weighted, and PD, each capturing distinct tissue characteristics that, taken together, support accurate lesion characterization, segmentation, and longitudinal follow-up \cite{tae2025current,alex2017semisupervised}. In practice, however, obtaining a complete set of sequences is often infeasible: long acquisition times increase patient discomfort and susceptibility to motion artifacts, contrast-enhanced scans raise safety and cost concerns, and retrospective or multi-site datasets frequently suffer from missing or corrupted modalities. These limitations have motivated growing interest in MRI image translation, which aims to synthesize a target modality from one or more available source modalities, thereby completing missing information without additional scanning.

Deep generative models have been central to progress in this area. Early efforts were dominated by generative adversarial networks (GANs), whose adversarial training enabled sharp, visually realistic synthesis and established strong baselines for both paired and unpaired translation. As the representative method for paired data, conditional models such as pix2pix \cite{isola2017image} use a source-conditioned generator to learn a direct mapping from source to target images. For unpaired settings, CycleGAN \cite{zhu2017unpaired} introduces a cycle-consistency constraint to align the two modalities without pixel-wise correspondence. These methods established strong baselines and remain widely used points of comparison for medical image synthesis.

However, GAN-based approaches are constrained by several well-known limitations. The adversarial min–max objective is difficult to optimize, resulting in unstable training and mode collapse \cite{goodfellow2020generative}; more critically in a clinical setting, GANs are prone to hallucinating anatomically plausible yet non-existent structures that are not supported by the source image \cite{wang2025generative}. Such fabricated details pose a serious risk when synthesized images inform downstream diagnosis or treatment planning.

These limitations motivate a reconsideration not only of the training objective but also of the backbone architecture on which such generators are built. Convolutional networks operate over a fundamentally local receptive field and therefore struggle to enforce consistency across spatially distant regions of an image. The Transformer architecture offers a complementary inductive bias: through the self-attention mechanism, every spatial location interacts directly with every other, enabling the model to capture long-range dependencies and global anatomical context \cite{vaswani2017attention,dosovitskiy2020image}. Motivated by this property, a growing body of work has adapted Transformers to medical image translation and reconstruction \cite{dalmaz2022resvit}, where preserving globally coherent anatomy across the entire field of view is essential.

However, existing image translation methods primarily focus on learning the mapping between image intensities, while the underlying structural and semantic information of the image is often not explicitly modeled. This limitation is particularly important in medical image translation. For example, in inter-modality MRI translation, different MRI modalities exhibit substantially different intensity distributions, while sharing largely consistent anatomical structures. Therefore, successful translation requires not only preserving the modality-invariant structural and semantic information, but also effectively modeling and transferring the modality-specific appearance, such as intensity and contrast. Specifically, in hierarchical image reconstruction models, such as Unet \cite{ronneberger2015u}, deeper layers tend to capture more abstract anatomical information, whereas upper layers progressively conduct style or intensity transformation. Recent studies have explored vector quantization with learned codebooks to represent images using discrete latent representations, which have shown promising results in capturing meaningful semantic information for image generation \cite{van2017neural,zhu2025addressing}, while conditional feature modulation provides an effective mechanism for controlling image-specific style and appearance \cite{perez2018film,park2019semantic}. These complementary advances motivate us to explicitly model the two aspects of image translation at different hierarchical levels: a discrete bottleneck for preserving modality-invariant structural content, and higher-level representations for capturing and transferring modality-specific appearance.

In this paper, we propose a novel framework, S2T-Unet, for inter-modality MRI translation by rethinking how different levels of the Unet representation can be utilized. Specifically, we introduce vector quantization at the bottleneck of the encoder to preserve modality-invariant structural and semantic information, while using the decoded representation as a condition to guide modality-specific style transformation toward the target MRI modality. Our framework builds upon the existing Unet architecture and requires no additional hyperparameter tuning beyond that of the original vector quantization module. Extensive experiments on a publicly available multi-contrast MRI dataset demonstrate that our approach consistently outperforms state-of-the-art methods.

\section{Methods}
\begin{figure}
\includegraphics[width=\textwidth]{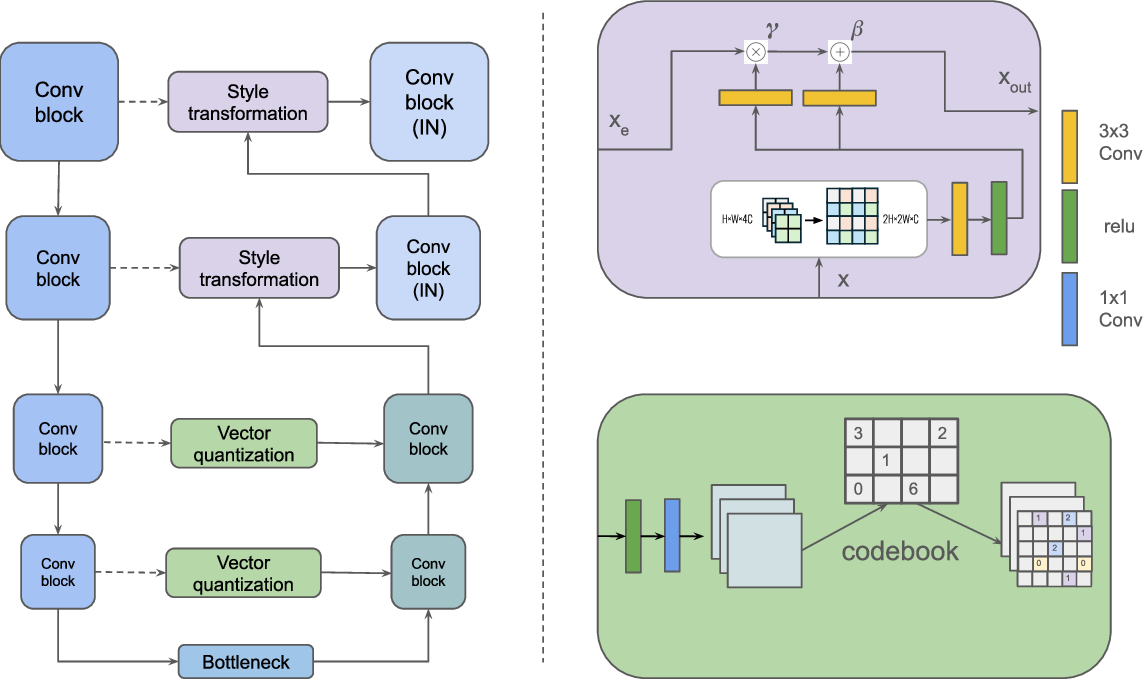}
\caption{Overview of S2T-Unet. It comprises three components: a Unet encoder–decoder, style transformation modules, and vector quantization modules. Dashed arrows denote features extracted before batch normalization and ReLU function, which are fed into the style transformation module. IN denotes that the normalization in the Conv block is instance normalization.} 
\label{fig1}
\end{figure}

The S2T-Unet architecture of our method is shown in Fig. \ref{fig1}. It consists of style transformation modules, vector quantization modules, and a Unet encoder-decoder structure. We assume in the MRI image translation tasks, the Unet decoder is responsible for the image translation or style translation, while encoder is responsible for information extraction, such as semantic information extraction. 

\subsection{Vector Quantization in the Low-Level Unet}
The key idea behind vector quantization is to represent an image using a finite set of discrete codes drawn from a learned codebook, a dictionary in which each entry captures a recurring structural pattern of the image. Instead of encoding an image as continuous, unconstrained features, a vector quantization model maps each local feature vector to its nearest entry in the codebook, yielding a compact discrete representation. Because the codebook is shared and its entries are repeatedly reused across images, this representation is encouraged to capture the essential anatomical structure rather than modality-specific appearance or noise.

This property is particularly well suited to intra-modality MRI translation. When a T1-weighted image is translated to T2, the underlying anatomy must remain unchanged. Since the skip connection of the lower-level U-Net carries the structural content shared between the source and target contrasts, we quantize it (Fig.~\ref{fig1}, bottom right) so that these structural features are passed to the decoder in a discrete form and reused during synthesis. By anchoring synthesis to a fixed vocabulary of learned structures, vector quantization constrains the model from fabricating anatomically implausible details, directly mitigating the hallucination problem of GAN-based approaches.

Formally, given an encoder feature map $z_{e}(x) \in \mathcal{R}^{h \times w \times C}$, where $C$ is channel numbers, and a codebook $E=\{e_k\}^{K}_{k=1}$, where $k$ is the code numbers. Vector quantization replaces each spatial feature of MRI by its nearest entry, $z_{q}(x)_{i,j}=e_k$ with $k=\arg\min_{m}||z_e(x)_{i,j}-e_m||_2$. Then the quantized latent space will feed into decoder and be concatenated with decoder features.

\subsection{Modality Transformation}
While vector quantization preserves the anatomical structure of the MRI image, its discrete nature inevitably discards continuous appearance information such as texture and contrast, which is precisely the modality-specific content that a translation model must alter. Moreover, the Unet decoder reconstructs the transformed image through direct feature concatenation, which may oversimplify the interaction between transformed representations and low-level source features, potentially limiting the exploitation of causal information. To recover this information while performing the translation, we draw on the idea of style transformation: the decoder features, which encode the target-modality appearance, serve as a conditioning signal, while the quantized skip-connection features from the encoder are treated as the content whose modality is to be transformed. Concretely, we design a Modality Transformation module at upper layers based on Feature-wise Linear Modulation (FiLM)~\cite{perez2018film}, which modulates the encoder features conditioned on the decoder features, thereby transferring the modality and compensating for the information loss introduced by vector quantization at lower level layers. 

In practice, the feature $\mathbf{x}^i$ denotes activation of $i$-th layer in decoder, and the height and width are $H^{i}$ and $W^{i}$, respectively. The feature from encoder is denoted by $\mathbf{x}_e$, and the height and width are $2H^{i}$ and $2W^{i}$, respectively. The upper-right part of Fig. \ref{fig1} illustrates the design. It can be formulated as below:
\begin{align}
    \mathbf{x}_{out}^i=(1+\gamma^i) \mathbf{x}_e^i +\beta^i \nonumber\\
    \gamma^i = f_{\gamma}(ReLU(f_c(f_r(\mathbf{x}^i)))) \nonumber \\
    \beta^i = f_{\beta}(ReLU(f_c(f_r(\mathbf{x}^i)))) 
\end{align}

where, $\gamma^i$ denotes scaling feature and $\beta$ denotes bias feature.
$f_r$ denotes the reshape function. $f_c$, $f_{\gamma^i}$ and $f_{\beta^i}$ are $3 \times 3$ conv layer.

\subsection{Objective Function}
To train the network, we use L1 norm to calculate the distance between the output and target images, $f(x)$ and $y$, respectively. The function is shown as:
\begin{equation}
    L_1=E_{x,y}||y-f(x)||_1
\end{equation}
To update the codebook in vector quantization module, we have applied vector quantization loss and commitment loss from this study \cite{van2017neural}. Thus, the whole loss function is:
\begin{equation}
    L=L_1+||sg[z_e(x)]-e_m||^2_2+\alpha||z_e(x)-sg[e_m]||^2_2
\end{equation}
where, $sg$ is a stopgradient operator. $\alpha$ is a coefficient to balance vector quantization loss, second term and commitment loss, third term. We use $\alpha=0.25$ here.

\section{Experimental Results}
We have evaluated our proposed network S2T-Unet on the IXI dataset \footnote{https://brain-development.org/ixi-dataset/}. The IXI dataset has 581 subjects. Each one includes T1-, T2- and PD-weighted MRI image. The voxel size of the MRI images is ($0.94 \times 0.94 \times 1.2~\mathrm{mm}^3$). We used 115 subjects for training and testing separately. The images are preprocessed by ants Python package, and HD-BET is used for skull stripping. We have conducted 4 one-to-one image translation experiments: T1 $\rightarrow$ T2, T2 $\rightarrow$ T1, T1 $\rightarrow$ PD, PD $\rightarrow$ T1. We use a batch size 32, learning rate of $3e-4$ for training. The models are trained with 20 epochs. Three metrics are used for the evaluation: Peak Signal-to-Noise Ratio (PSNR), Structural Similarity Index Measure (SSIM), and Root Mean Squared Error (RMSE).

We compare our method with popular CNN-based and transformer-based methods: Unet \cite{ronneberger2015u}, CycleGAN \cite{zhu2017unpaired}, Pix2Pix \cite{isola2017image}, and ResViT \cite{dalmaz2022resvit}. The quantitative performance is shown in Table \ref{tab1}. Our method achieves competitive or superior performance across all four translation tasks compared with both CNN-based and Transformer-based methods. In particular, it consistently outperforms the Transformer-based ResViT in terms of PSNR and RMSE, while achieving comparable or slightly improved SSIM scores. Notably, these results are obtained using a purely CNN-based architecture built upon the Unet framework, without introducing adversarial loss. Compared with Unet, our base model, the proposed approach substantially improves the translation performance across all four tasks. This improvement suggests that the encoder can effectively preserve and pass semantic and structural information to the decoder, thereby reducing structural information loss during reconstruction. Moreover, the proposed approach consistently achieves higher PSNR than the Transformer-based ResViT across all four translation tasks. This performance advantage suggests that the proposed modality transformation module effectively leverages decoder features to modulate encoder features, facilitating the synthesis of target-modality intensity and contrast.


\begin{table}
\caption{Quantitative performance comparison between CNN and Transformer-based architectures for one-to-one image-to-image translation using the IXI dataset. }\label{tab1}
\begin{tabular}{l|ccc|ccc|ccc|ccc}
\hline
\multirow{2}{*}{Model} & \multicolumn{3}{c|}{T1$\rightarrow$T2} & \multicolumn{3}{c|}{T2$\rightarrow$T1} & \multicolumn{3}{c|}{T1$\rightarrow$PD} & \multicolumn{3}{c}{PD$\rightarrow$T1} \\
\cline{2-13}
 & PSNR & SSIM & RMSE & PSNR & SSIM & RMSE & PSNR & SSIM & RMSE & PSNR & SSIM & RMSE \\
\hline
UNet     & 24.95 & 0.885 & 0.0576 & 21.81 & 0.862 & 0.0846 & 22.63 & 0.815 & 0.0752 & 21.63 & 0.813 & 0.085 \\
Pix2Pix  & 26.50 & 0.899 & 0.0477 & 23.38 & 0.875 & 0.0693 & 25.02 & 0.891 & 0.0572 & 23.30 & 0.874 & 0.07 \\
CycleGAN & 25.44 & 0.889 & 0.0522 & 22.69 & 0.869 & 0.075 & 23.45 & 0.858 & 0.066 & 22.51 & 0.847 & 0.076 \\
ResViT   & 27.37 & 0.917 & 0.0432 & 23.72 & 0.887 & 0.0681 & 25.11 & 0.899 & 0.0569 & 23.28 & 0.886 & 0.0719 \\
Ours     & 27.48 & 0.916 & 0.0428 & 24.10 & 0.893 & 0.0657 & 25.34 & 0.900 & 0.0554 & 23.78 & 0.892 & 0.0677 \\
\hline
\end{tabular}
\end{table}

\begin{figure}
\centering
 \begin{subfigure}{0.48\textwidth}
        \centering
        \includegraphics[width=\textwidth]{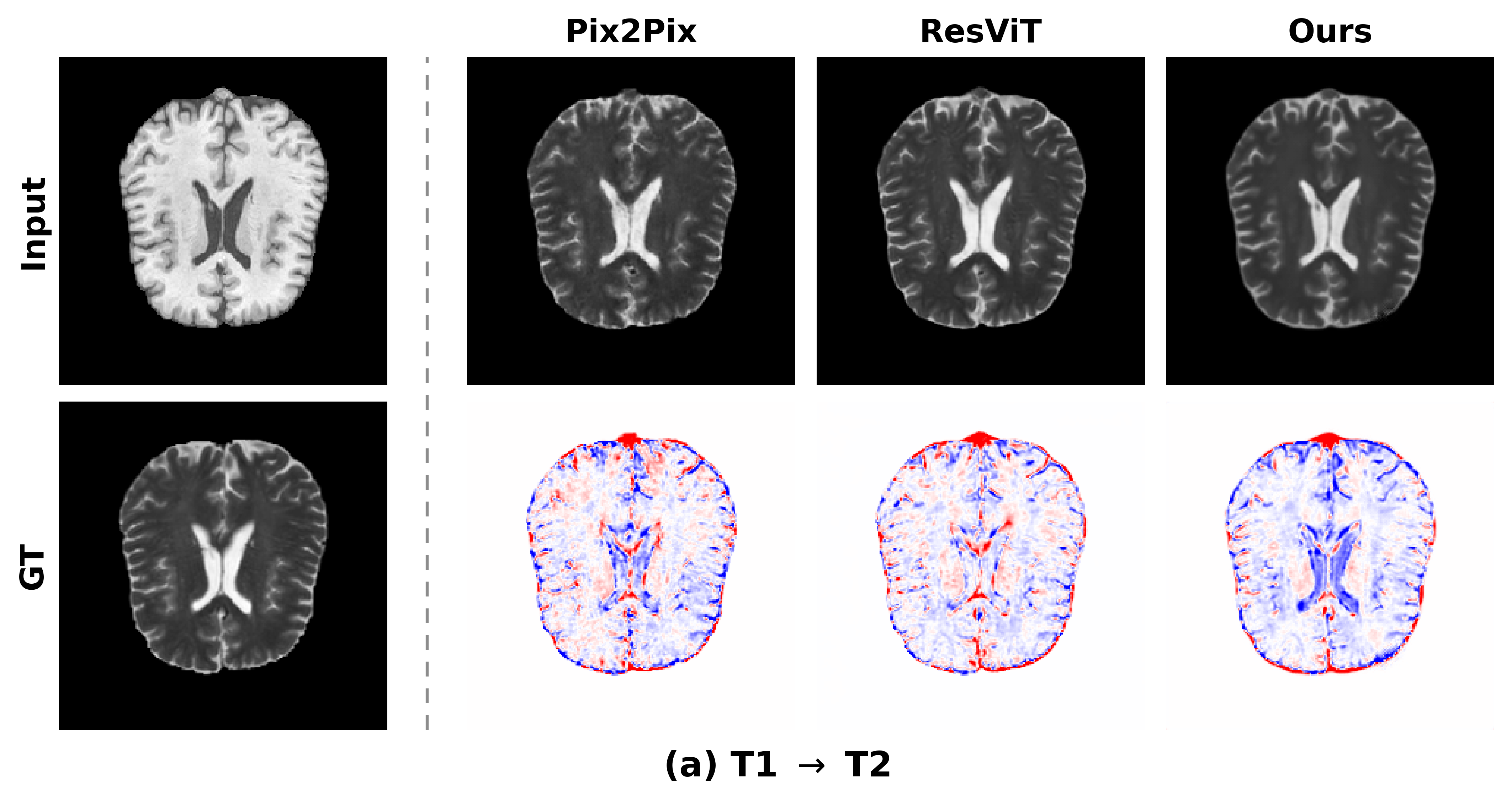}
    \end{subfigure}
    \hfill
 \begin{subfigure}{0.48\textwidth}
        \centering
        \includegraphics[width=\textwidth]{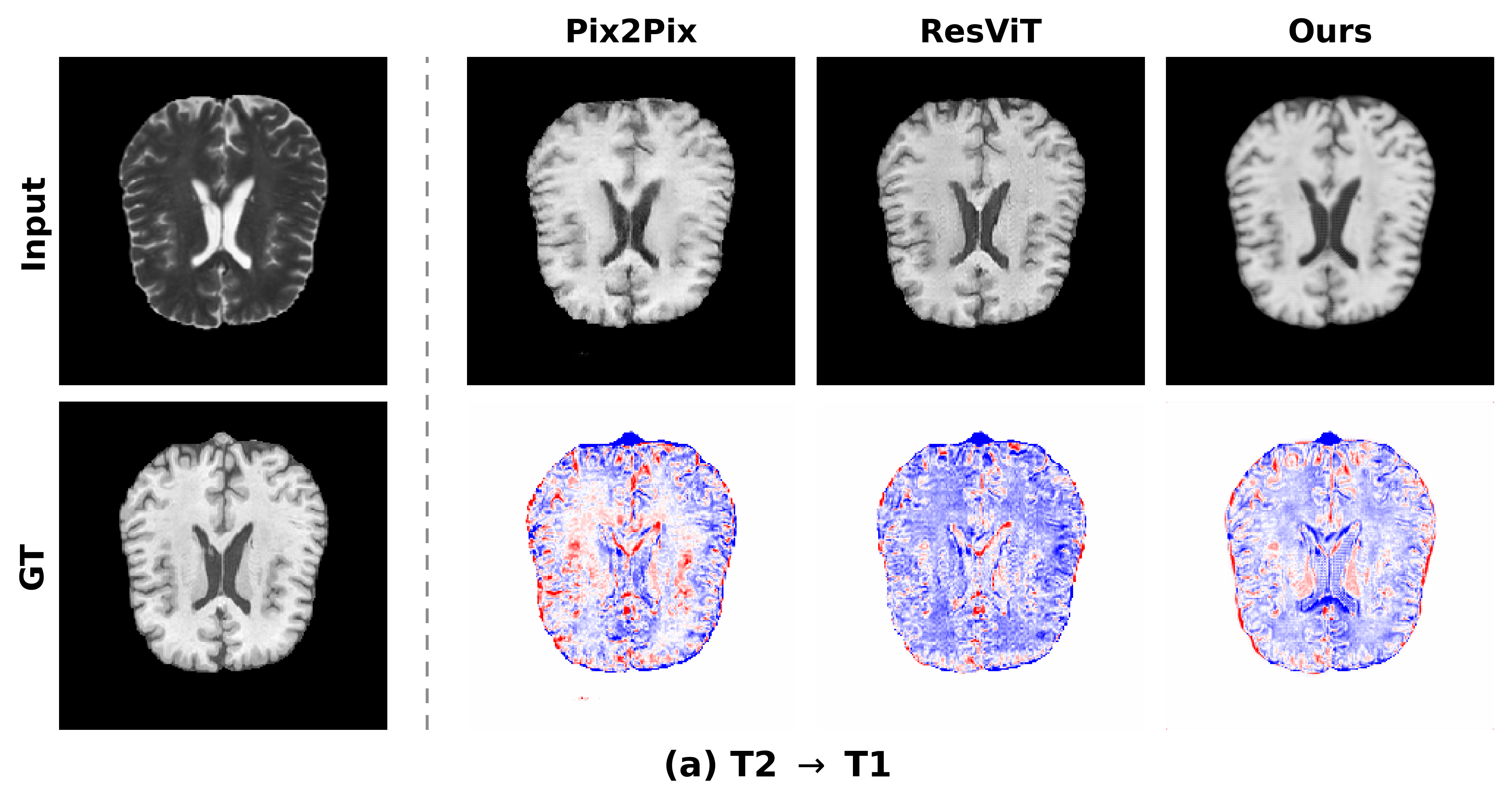}
    \end{subfigure}
    \hfill
\begin{subfigure}{0.48\textwidth}
        \centering
        \includegraphics[width=\textwidth]{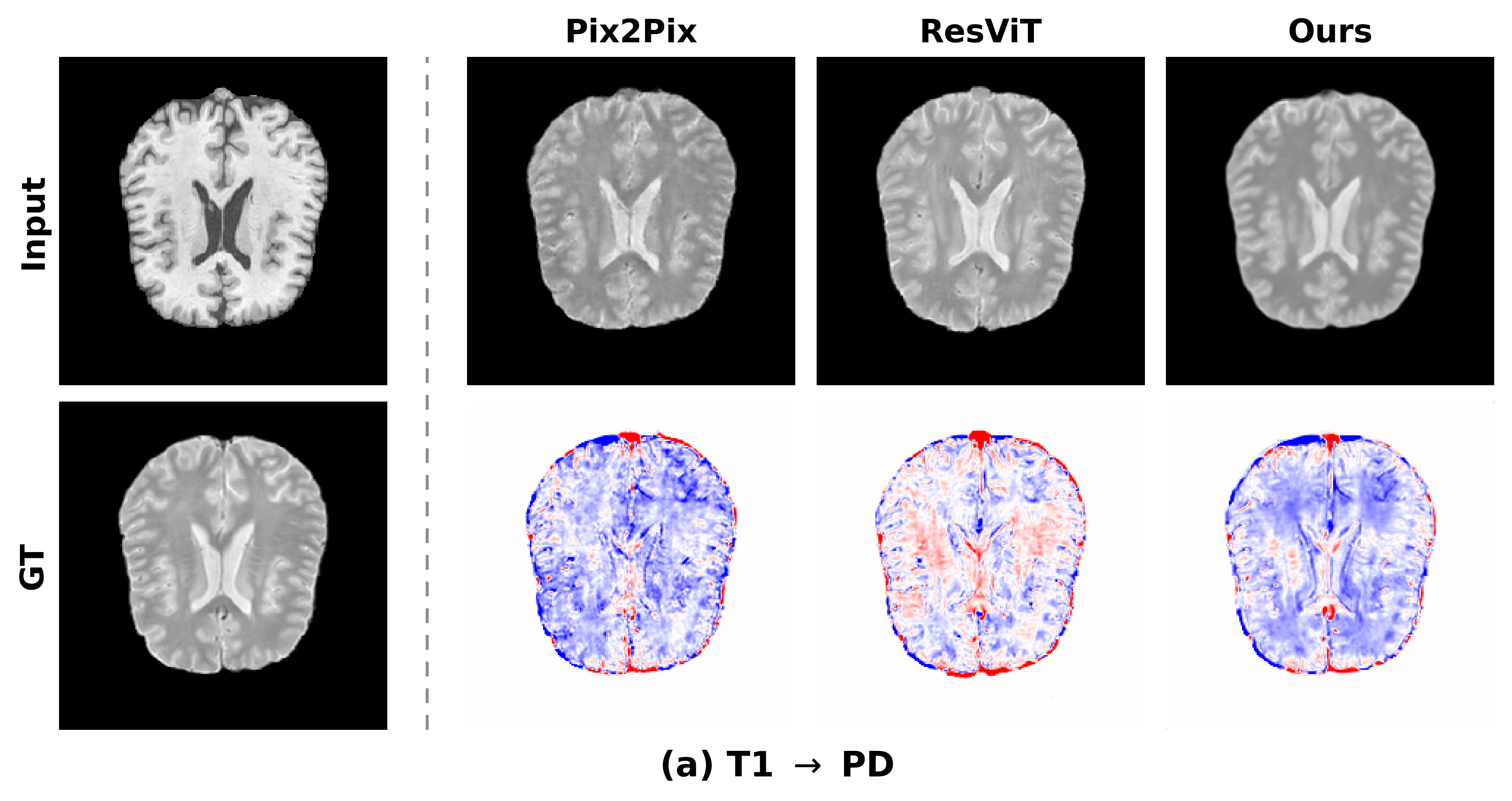}
    \end{subfigure}
    \hfill
\begin{subfigure}{0.48\textwidth}
        \centering
        \includegraphics[width=\textwidth]{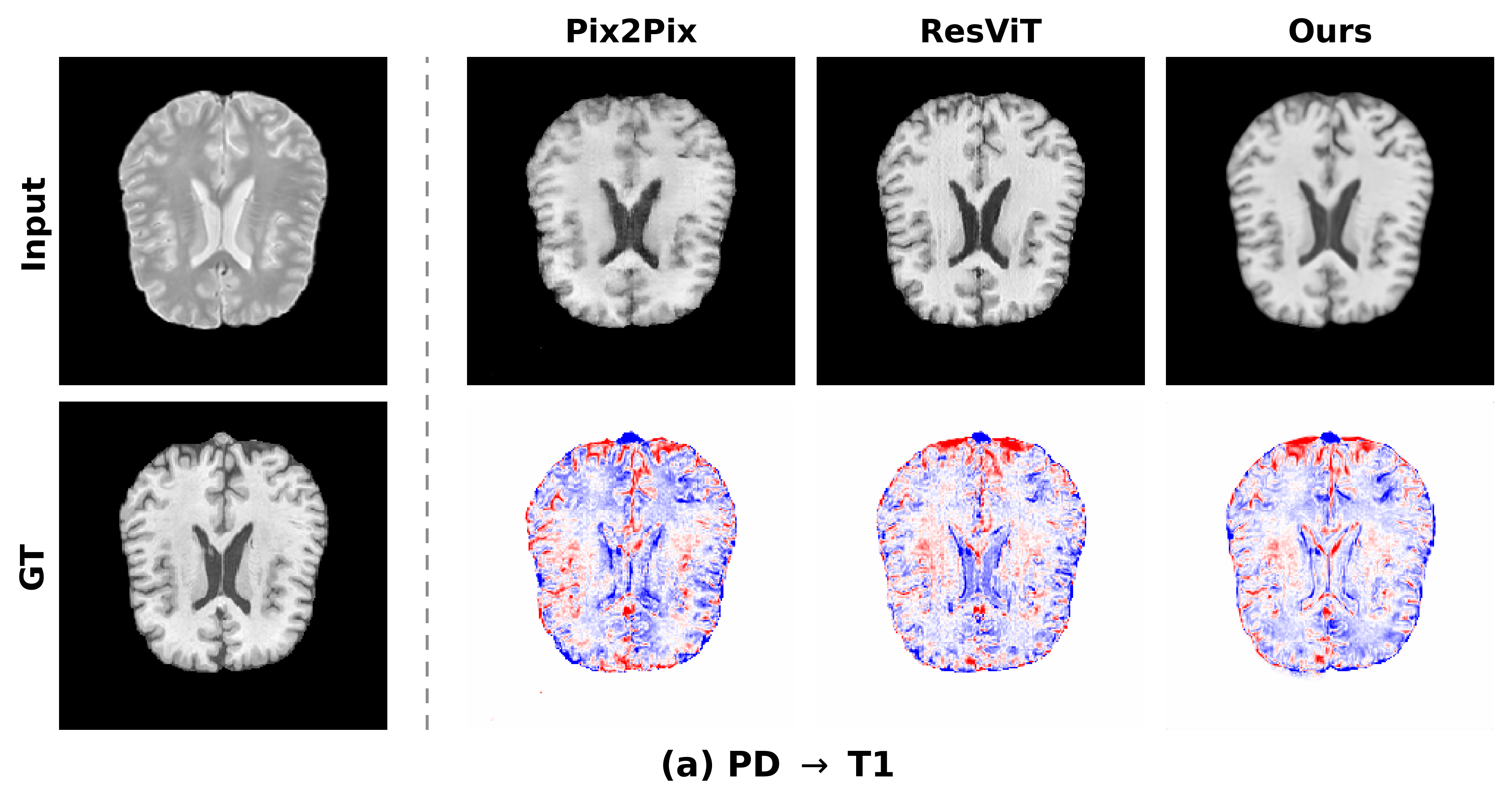}
    \end{subfigure}
\caption{Qualitative comparison of MRI cross-modality translation in four directions: (a) T1 $\rightarrow$ T2, (b) T2 $\rightarrow$ T1, (c) T1 $\rightarrow$ PD, and (d) PD $\rightarrow$ T1. In each panel, the input (source modality) and ground truth (GT, target modality) are shown on the left, and the synthesized results of Pix2Pix, ResViT, and our method are shown on the right. The bottom row of each panel shows the corresponding difference maps between the synthesized image and the GT, where red and blue indicate positive and negative errors, respectively.} 
\label{fig2}
\end{figure}

We visualize our synthesized MRI modality image in Fig. \ref{fig2}. Overall, the proposed method produces synthesized images with appearance characteristics that are visually closer to the target modality while maintaining the anatomical structures of the source image. Notably, the difference maps reveal a clear contrast in error structure between our method and the competing baselines. The residuals produced by our method are spatially coherent and concentrated at genuine anatomical boundaries, whereas the other methods exhibit noisy, disorganized difference maps in which positive and negative errors (red and blue) are densely intermixed throughout the image. We attribute this to the effect of adversarial training: adversarial losses encourage the generator to hallucinate realistic-looking high-frequency texture, but such texture is inherently stochastic and rarely aligns pixel-wise with the ground truth, resulting in scattered, high-frequency error patterns. In contrast, our deterministic formulation, trained without any adversarial objective, avoids fabricating such texture and instead yields errors that are spatially consistent and largely confined to regions of genuine structural mismatch.

\begin{table}[]
    \centering
    \begin{tabular}{c|ccc|ccc|ccc}
    \hline
        \multirow{2}{*}{} & \multicolumn{3}{c|}{without VQ} & \multicolumn{3}{c}{without MT}& \multicolumn{3}{|c}{with both} \\
        \cline{2-10}
         & PSNR & SSIM & RMSE & PSNR & SSIM & RMSE& PSNR & SSIM & RMSE\\
         \hline
        S2T-Unet & 26.10 & 0.900 & 0.0510 & 25.20 & 0.888 & 0.0559& 27.48 & 0.916 & 0.0428  \\
        \hline
    \end{tabular}
    \caption{Ablation study for vector quantization (VQ) and modality transformation (MT). }
    \label{ablation}
\end{table}

Table \ref{ablation} presents the ablation study of the vector quantization and modality transformation modules. Removing either vector quantization or modality transformation leads to a noticeable degradation in performance, confirming that both components contribute to the overall effectiveness of S2T-Unet. In particular, removing modality transformation results in a larger performance drop, with the PSNR decreasing from 27.48 dB to 25.20 dB, compared with a decrease to 26.10 dB when vector quantization is removed. This suggests that the modality transformation module plays a more direct role in the reconstruction process, as it explicitly modulates the encoder features using decoder features to adapt the representation to the target modality. In contrast, vector quantization is mainly introduced to preserve modality-invariant structural and semantic information at the bottleneck, and its contribution may therefore be less directly reflected in the final reconstruction. Nevertheless, the removal of vector quantization still causes a substantial performance degradation, indicating that structural information preserved by vector quantization is important for achieving the best translation performance when combined with modality transformation.

\section{Conclusion}
We proposed S2T-Unet, an MRI inter-modality image translation framework. The model learns semantic information, i.e., the modality-invariant feature, from a codebook, thereby avoiding changes to the underlying image structure. Vector quantization at the lower-level bottleneck anchors synthesis to a discrete vocabulary of structural codes, preserving modality-invariant anatomy and mitigating the hallucination of anatomically implausible details, while a modality transformation module at the upper levels uses decoder features to condition and transfer the target-modality appearance onto encoder representations. Experiments on the IXI dataset across four translation tasks show that our purely convolutional, deterministic formulation, requiring neither adversarial training nor Transformer-based self-attention, matches or surpasses popular GAN-based and Transformer-based baselines, indicating that explicit structure-appearance separation offers an effective and stable alternative to adversarial or attention-based designs. Future work includes extending the framework to 3D volumetric translation and evaluating generalization on additional multi-contrast and multi-modal datasets.

%

    

\begin{credits}
\subsubsection{\ackname} 
We are grateful for access to the University of Heidelberg’s IWR HPC service for running simulations used in this study.
\end{credits}

%
%
%
\bibliographystyle{splncs04}
\bibliography{ref}

\end{document}